\documentclass[11pt, titlepage]{article}
\usepackage{graphicx}

\usepackage[margin=1in]{geometry}
\usepackage{amssymb}
\usepackage{amsmath}
\usepackage{tikz-cd}

\usepackage{hyperref}
\usepackage{mathtools}
\usepackage{shuffle}
\usepackage{caption, copyrightbox}

\title{Coinductive guide to inductive transformer heads}

\newcommand{\id}{\mathrm{id}}

\begin{document}

\begin{minipage}[h]{\textwidth}
    \date{}
    \author{Adam Nemecek \\ adam@cofunctional.ai}

    \maketitle

\end{minipage}

\section{Abstract}

\textbf{We argue that all building blocks of transformer models can be expressed with a single concept: combinatorial Hopf algebra}.

Transformer learning emerges as a result of the subtle interplay between the algebraic and coalgebraic operations of the combinatorial Hopf algebra. Viewed through this lens, the transformer model becomes a linear time-invariant system where the attention mechanism computes a generalized convolution transform and the residual stream serves as a unit impulse.

Attention-only transformers then learn by enforcing an invariant between these two paths. We call this invariant \textbf{Hopf coherence}. Due to this, with a degree of poetic license, one could call combinatorial Hopf algebras "tensors with a built-in loss function gradient". This loss function gradient occurs within the single layers and no backward pass is needed. This is in stark contrast to automatic differentiation which happens across the whole graph and needs a explicit backward pass. This property is the result of the fact that combinatorial Hopf algebras have the surprising property of calculating eigenvalues by repeated squaring.

\section{Introduction}

With the rise of popularity of transformer models \cite{Vaswani2017}, there have been concerns about the interpretability, or lack thereof, of said models. Such concerns are warranted, transformer models exhibit behaviors that appear surprising. 

We analyze these behaviors in the context of combinatorial Hopf algebra, which is a tensorial bialgebra. This algebra turns out to be uniquely suited for interpretation and understanding of transformer models and, as we show in future work, machine learning models in general. 

Despite the fact that this is, to our knowledge, the first paper which interprets machine learning in terms of Hopf algebras, we are confident that this venue of research will prove to be a quintessential semantic bridge between the representation of machine learning models in memory as IEEE 754 and the observable behaviors of said models. 

The paper is structured as follows:
section 3 discusses the idea of duality in general and briefly discusses how it has been used.
Section 4 is an introduction to Hopf algebras, and section 5 extends that to combinatorial Hopf algebra.
Section 6 discusses how these concepts apply to transformer models.

\section{Duality}

\subsection{Induction}

\textbf{Induction}, the basic idea behind algebra and recursion, is a well-understood concept in both mathematics and computer science. Fundamentally, induction is about building up objects from smaller components via the use of a \textbf{product} operation.

$$ C \otimes C = C $$

Inductive data structures are defined in terms of their \textbf{constructors}

$$ nil: 1 \rightarrow U $$
$$ cons: D \times U \rightarrow U$$

where $U$ is the the data type being defined. One can think of the linked list being formalized as:

$$ \zeta: 1 + D \times U \rightarrow U $$

\cite{Barbosa2022}

\subsection{Coinduction}

\begin{quote}
  
"[...] coalgebras are about observation. We can think of a coalgebra $f: X \rightarrow 1 + AX$ as observing about an entity whether it contains something $A$-detectable or not, and if so which element of $A$ it detects. Having observed something it modifies it. The final coalgebra has as elements all possible outcomes of the behavior you might observe. Do you still have observations to add as list elements? If ever no, we have a finite list. If always yes, we have an infinite list. And there’s no other behavior that can be detected."
  \cite{Corfield2011}
\end{quote}

\textbf{Coinduction}, a dual of induction, has been relegated to the realm of computer science esoterica. Coinduction, and the related coalgebra, is about analysis, namely specifying how things break down by defining \textbf{coproduct}:

$$ C = C \otimes C $$

Turning an algebra into a coalgebra is as simple as reversing the arrows. Using this principle, we can create \textbf{destructors} from linked list constructors above like so:

$$ head: U \rightarrow 1 $$
$$ tail: U \rightarrow D \times U $$

The coinductive list formula is then:
$$ \alpha: U \rightarrow 1 + D \times U $$.

\cite{Barbosa2022}

Coinduction is fundamentally about defining how something changes in response to observation.

As such, it is the underlying principle of recurrent relationships, generating functions, dynamic systems, coinductive data structures, and fundamentally anything that deals with modeling of observed, possibly infinite, behavior.

Here are some examples of coinduction:

\subsubsection{Recurrence relations}
\textbf{Recurrence relations} is a way of defining terms of a sequence in terms of combinations of previous elements of the sequence.

The connection between Hopf algebras and recurrence relations is discussed in \cite{Peterson1980}.

\subsubsection{Generating function} 

\begin{quote}
A generating function is a device somewhat similar to a bag. Instead of carrying many little objects detachedly, which could be embarrassing, we put them all in a bag, and then we have only one object to carry, the bag.

George P\'olya, Mathematics and plausible reasoning
\end{quote}

\textbf{Generating function} is a generalization of formal power series, itself a generalization of power series. 

Ordinary generating function are functions of this format:
\[  G(a_n; x) = \sum_{n=0}^{\infty} a_n x^n  \]

The nomenclature is not accurate, \textbf{generating functions are not functions} per se, they provide a way of expressing something in terms of a sum of lower dimensional elements with additional constraints on what combines with what via the use of "powers". However, these powers are indeterminate, one is not expected to substitute for $x$ and evaluate. Generating functions are for handling infinite sums and establishing recurrences. As such, they have been used extensively in the context of analytic and enumerative combinatorics \cite{Flajolet2009}.

The powerful idea of generating functions is that they provide a unified interface between linearity and non-linearity due to the fact that generating functions provide a stream of semi-rings which can be further combined to a single generating function producing semi-rings.

As we will see, the this idea is closely related to the idea of coproduct in Hopf algebras.

In the context of machine learning, automatic differentiation can be seen as an an instance of generating functions \cite{Carothers2012}.

\subsubsection{Stream algebra}
We very briefly mention streams, or infinite lists, as they provide a good mental model for how to think about coalgebraic programming. Stream algebras are closely related to generating functions, and have been studied as a way of modeling recurrences, streaming automata, process algebras etc. One can think of them as "streaming representation-changers" \cite{Gibbons2004} as they change their internal state in response to external changes.

One way of working with defining coalgebraic object is in terms of (often self-referential) streams or infinite lists by specifying an initial value and an update function. Due to the requirement for laziness, the Haskell programming language is a natural choice for this programming paradigm. 

By self-referential we mean that the stream we are defining appears on both sides of the assignment operator.

Consider this Haskell definition a stream representing the natural numbers:

$$ nats = 1 : map(+1) nats $$

Note how $nats$ appears on both sides of the assignment operator and how it is defined in terms of an initial state and an update rule. It is for this reason that \cite{Rutten2003} calls coinductive streams \textbf{differential equations of programming}. \cite{Clenaghan2018} provides a complete Haskell implementation of such programming paradigm. Said paper praises the coinductive approach for it s "economy of statement and notation, whilst embracing variety of approach".

The fundamental difference between recursion and corecursion is that recursive Fibonacci function returns only a single value,
while corecursive fibonacci returns a \textbf{stream} of all Fibonacci numbers. 

The stream algebra has a close relationship with Hopf algebra \cite{Zanasi2014}.

\subsubsection{Dynamic programming}

Dynamic programming can be naturally expressed in terms of recurrences. 
The Bellman equation 

$$ V(s) = max_a(R(s, a) + \gamma V(s')) $$

is self-referential. The relationship between coalgebras and dynamic programming is discussed for example in \cite{Hinze2015}.

\subsection{Applications}

Coalgebras and coinduction have been slowly but surely gaining popularity.

This approach of building up things from their constituent parts enables reasoning about all possible paths and interactions in a black-box dynamic system by analyzing all the possible interdependent interactions among the single atoms starting from the bottom \cite{Barbosa2022}. The field of bisimulation defines equivalence in terms of observation equivalences \cite{Sangiorgi2012} and thereby verifies the dynamic behaviors of systems.

\begin{itemize}
	\item \cite{Vajjha2021} has applied coinduction in the context of verified reinforcement learning.
	\item \cite{Hur2013} discusses how coinduction allows for building proofs incremental by combining small proofs into larger proofs.
	\item \cite{Nguyen2022} discusses backpropagation as a coalgebra.
	\item \cite{Mastorou2022} extend Haskell to verify coinductive proofs.
	\item \cite{Pous2016} discusses how coinduction enabled validating global properties by checking only local properties.
	\item \cite{Kozen2017} introduced \textit{CoCaml} which discusses the idea of coinductive programming as programming with a coinductive equation solver.
\end{itemize}

\section{Hopf algebra}

\textbf{Hopf algebra} is a tensor bialgebra $A$ over a field $C$ meaning it is both a tensor algebra and a tensor coalgebra at once.

In the diagram below, we refer to the counit-unit path as "unit impulse path" and the top and bottom paths as the "convolution paths".

Hopf algebra enforces coherence between the two paths by updating internal state in response to input \cite{Ahman2014}.
This is a very powerful invariant for reasoning about the behavior of a linear time-invariant system \cite{Zanasi2014}.

\[
\begin{tikzcd}[row sep=3.6em,column sep=1em]
& A\otimes A
	\arrow[rr,"S\otimes\id"] && A\otimes A 
	\arrow[dr,"m"] \\
A \arrow[ur,"\Delta"] \arrow[rr,"\epsilon (counit/trace)"] \arrow[dr,"\Delta"'] && C \arrow[rr,"u (unit/cotrace)"] && A \\
& A \otimes A \arrow[rr,"\id\otimes S"'] && A\otimes A \arrow[ur,"m"']
\end{tikzcd}
\]

Considering popularity of Hopf algebra in other scientific fields \cite{Hazewinkel2004} it was only a matter of time before they are used in machine learning.

Hopf algebra is defined by the following operations.

\begin{itemize}
	\item unit: $u$: $C \rightarrow A$
	\item product: $m$: $A \otimes A \rightarrow A$

	\item counit: $\epsilon: A \rightarrow C$
	\item coproduct: $\Delta: A \rightarrow A \otimes A $

	\item antipode: $S: A \rightarrow A$
\end{itemize}

\subsection{Unit}

The definition of unit is straightforward, it takes an element of $C$ and construct and element of $A$:

$$u: C \rightarrow A$$

and obeys the rule

$$ u \cdot e = e = e \cdot u $$ for all elements of $A$.

One should however think of the unit in relationship to the equalizer.

\subsection{Product}

The product is the standard tensor product 
$$ (a \otimes c) (b\otimes d)= (ab \otimes cd) $$

\subsection{Counit}

Counit is similar to the idea of trace from linear algebra and as such provides a feedback loop operator from control theory \cite{Hasegawa2022}.

It is a dual of the unit and as such it works as a coequalizer.

\subsection{Coproduct}

The coproduct provides a way of generating all possible splittings of a subset into disjoint pieces \cite{Majid1995}.

$$ \Delta(x^n) = \sum _{i=0}^n x^i \otimes x^{n-i} $$ where $x^0= 1$.

One can think of it as converting something into its generating function \cite{Hazewinkel2005} or into a sum of elements of lower graded subcoalgebra such as simplices.

In the context of physics, it is a probability density function, a total probability mass that's being shared out among different spaces \cite{Majid1995}.

\begin{equation*}
	\begin{aligned}
\Delta(wz) & = \Delta(w)\Delta(z) \\
\Delta(1) &= 1 \otimes 1 \\
\Delta(x) &= 1\otimes x + x \otimes 1 \\
\Delta(x^2) &= 1 \otimes x + x \otimes x + x \otimes 1
    \end{aligned}
    \label{equ:ho}
\end{equation*}

For a more complicated example:

\begin{equation*}
	\begin{aligned}
	(id \otimes \Delta)(\Delta(x^2)) & =   \\
	  & = (id\otimes \Delta)(1 \otimes x^2 + x \otimes x + x^2 \otimes 1) \\
  	  & = 1 \otimes \Delta(x^2) + x \otimes \Delta(x) + x^2 \otimes \Delta(1) \\
  	  & = 1 \otimes ( 1 \otimes x^2 + x \otimes x + x^2 \otimes 1) + x \otimes (1 \otimes x + x \otimes 1) + x^2 \otimes (1 \otimes 1) \\
  	  & = 1 \otimes 1 \otimes x^2 + 1 \otimes x \otimes x + 1 \otimes x^2 \otimes 1 + x \otimes 1 \otimes x + x \otimes x \otimes 1 + x^2 \otimes 1 \otimes 1
    \end{aligned}
    \label{equ:ho}
\end{equation*}

Future work will discuss how superposition emerges out of an extension of a product and coproduct, namely the phased biproduct \cite{Tull2020}. But fundamentally, this arises out of the fact that when multiplying a sequence of Hopf algebras, one can select which ones to multiply and as a result, the coproduct represents a combinatorial object which includes all possible ways to choose \cite{Diaconis2012}.


\subsection{Antipode}

Since not all transformations are invertible, a weaker structure, \textbf{antipode}, provides a nonlocal "linearized inverse". Antipode doesn't provide an inverse for single elements but for linear combinations.
One can think of the antipode as a complex conjugate. However, if the Hopf algebra is finite dimensional, then the antipode is an inverse \cite{Majid1995}.

The antipode is a defined as:

$$ S \cdot S = id $$

$$ S(hg) = S(g)S(g) $$

The antipode serves as an intertwining operator, namely a equivariant linear map between two representations which gets updated as the Hopf algebra "learns".

In the context of interacting particle systems, the intertwiner provides a symmetric exclusion process \cite{Redig2018}.

The importance of the antipode will become apparent in the context of Hopf convolution. 

\subsection{Unit impulse, Convolution \& Hopf coherence}

Given the algebraic operations of the Hopf algebra, one can define the unit impulse $\delta$, also known as Dirac delta, as:

$$ \delta = \epsilon * u $$

\cite{Christiansen2022} discusses how traces of evolution operators can be evaluated as integrals over Dirac delta functions.

The purpose of the unit impulse is to serve as multiplicative identity for convolution $\ast$ which is defined as:

$$ (f \ast g) = m (f \otimes g) \Delta $$

The fundamental advantage of this convolution over the standard one is, as is the custom with bi-algebras, it provides a way of controlling how splitting-up and recombination works via the coproduct and product.

The unit impulse delta function then provides the multiplicative identity:
$$ f \ast \delta = f $$

Arguably the most important property of Hopf algebra is the fact that this invariant has to hold:

$$ m(id \otimes S) \Delta = \epsilon * u $$

We refer to this invariant as the \textbf{Hopf coherence}.

Hopf coherence enforces that the algebra updates its internal state in order for the unit impulse path to be equal the convolution path. 

The antipode plays an important part in enforcing this. The antipode reconstructs itself in order for this coherence to hold. This reconstruction process is a result of the "Tannaka--Krein duality" or "Tanaka reconstruction theorem" \cite{ncatlabTannaka} \cite{Pareigis1994}. The details of this reconstruction process as well as a proof thereof is provided in \cite{Majid1995}.

 The exact antipode reconstruction algorithm has been a subject of much research for example in \cite{Berlin2019}.

\section{Combinatorial Hopf algebra}

Combinatorial Hopf algebra is a Hopf algebra where the coproduct is defined as the \textbf{shuffle product}. Besides combinatorics \cite{Grinberg2014}, this algebra has also been explored in the context of control theory \cite{DuffautEspinosa2014}. For a classical introduction into the material, consider \cite{Rota1979}. 

\subsection{Shuffle product}

Originally, the shuffle product arose in the context of card shuffling.

In the setting of non-commutative algebras setting, the shuffle product provides a way of generating all the ways in which two words can be interwoven. For another example, imagine a situation where cars from two lanes merge into one lane. The shuffle product is a combinatorial object that represents all the possible ways in which cars from the two lanes can be interleaved to merge into one. Since word composition is non-commutative, the shuffle product is a natural match for representing combinatorial objects of words.

For example:

\begin{equation*}
	\begin{aligned}
		ab \shuffle ab &= 4aabb + 2abab \\
		ab \shuffle ba &= abab + 2abba + 2baab + baba
    \end{aligned}
    \label{equ:ho}
\end{equation*}

\cite{Lothaire1997}

\textbf{The most surprising property of the combinatorial Hopf algebra is that one can calculate the eigenvectors by repeated squaring (coproduct-product)} \\
\cite{Aguiar2013}.

 \cite{Pang2014} provides some intuition into this coproduct-product operation and how they correspond a repeated splitting and combination of combinatorial objects.

 \cite{Diaconis2012} uses this property in the context of diagonalization of Markov chains in natural bases.
 


The shuffle product also allows for interpreting differential equations combinatorially \\
\cite{Mishna2008}.

\section{Transformers}

Since first being introduced in \cite{Vaswani2017}, the transformer architecture has become one of the most widely studied architectures and as such has been discussed extensively for example by \cite{Phuong2022}, \cite{Elhage2021}. Due to this, we only concentrate on selected parts namely the attention mechanism and the residual stream.

In this section, we argue that the attention-only transformer model can be understood in terms of combinatorial Hopf algebras and as such can be analyzed as a \textbf{linear time-invariant system} \\ \cite{Espinosa2018}.

\textbf{The summary of our argument is that since the residual stream has no preferred basis, it should be understood as a trace/counit/unit impulse, and since the attention mechanism is a positive definite matrix, it should be understood as a transfer function of the system}.
\textbf{The model learns by enforcing Hopf coherence between the two paths.} 
 
The attention heads continuously update the residual stream (trace) and their internal state in order to enforce the invariance between the unit impulse path and Hopf coherence. 

Combinatorial Hopf algebra is a good match since it captures the non-commutativity of word composition.

\subsection{Residual stream}

The residual stream has been described as a channel used by single components of the model to communicate among themselves.

We argue that the \textbf{residual stream is the trace/counit/unit impulse} of the model. The major hint is that the residual stream has been observed to have \textbf{no privileged basis} \cite{Elhage2021}. This is analogous to the behavior of the trace in linear algebra where the trace of a matrix is also independent of the basis. While in linear algebra the trace is a scalar, \textbf{categorical trace is more akin to formal power series} \cite{Hines2007}.

As such the residual stream serves a similar purpose to the evaluation trace in automatic differentiation, namely establishing recurrences. This is unsurprising considering the fact that automatic differentiation is a recurrence relationship based on formal power series \cite{Carothers2012}, in particular the Taylor series \cite{Hoffman2014}. 
 
The residual stream is a recurrence relationship that attention heads use for both steering (by writing into it) and being steered by it (by reading from it). The values in the stream are continuously updated in response to input in order to enforce Hopf coherence.
 
The residual stream provides the notion of a feedback analogous to feedback in control theory or linear systems while the attention mechanism is the transfer function.

In the context of linear time-invariant systems, the transfer function is the inverse Laplace transform of the unit impulse response and vice-versa. 

This duality can be understood as the trace-transfer function duality \cite{Dold1985}.

It can also be understood as the duality between the trace and the fixed point \cite{Hasegawa2004} that's analogous to Tannaka--Krein duality. By interacting with the residual stream, attention heads are steering the direction of the fixed point search algorithm that the transformer performs.

Future research will investigate the exact nature of this interaction.

\subsection{Attention}

\textbf{The attention mechanism's role is the same as that of a transfer function in a linear time-invariant system, namely it calculates the frequency response of the transformer model}, in the case of transformers, the output token.

Our interpretation is based on the fact that attention matrices have been observed to have real positive eigenvalues. From that we conclude that they are symmetric positive definite matrices. This symmetry enables an input-output symmetry \cite{Majid1995}.

Attention mechanism operates on three inputs: \textbf{query}, \textbf{key}, \textbf{value} and generates one \textbf{output}. Please reference \cite{Vaswani2017} for explanation of these terms.

\cite{Elhage2021} formalizes the model as such:

\[  T =  Id \otimes W_UW_E + \sum_{h \in H } A^h \otimes (W_UW_O^hW_V^hW_E) \]
\[  A =  softmax(t^T \cdot W_E^TW_Q^TW_KW_E \cdot t) \]

Furthermore, they analyze the attention mechanism by splitting it into two circuits, \textit{QK (query-key)}, and \textit{OV (output-value)} and the associated matrices $W_{QK}$ and $W_{OV}$.

These two circuits roughly correspond to coproduct-product relationship of the convolutional path where the coproduct splits things and product recombines them as discussed in \cite{Diaconis2012} or \cite{Redig2018}.

\subsubsection{QK circuit}


The QK circuit assigns an attention score for a given query and key token. This score indicates how much the query wants to attend to the particular key token. We mostly agree with the interpretation of \cite{Chen2021} and \cite{Yuan2021} which understand this circuit as Markov chain transition matrices, as they are row-stochastic matrices, with each row summing to 1.

$$ \sum_{j=1}^{\alpha}P_{i,j} = 1 $$

This circuit fundamentally distributes the unit of attention among the potential candidates. Since it is a distributive operation, it is to be understood as a part of the coalgebraic circuit.

These weights can be understood as \textbf{stochastic recurrence relations}. They are kept around for as long as they are "interesting" and after that, they are gradually pruned.

Since the QK circuit generates the possible candidates, it can be understood as providing induced representation of the Markov chain. Induced representation is a tool in representation theory that enables building representations of large objects from representations of small objects. \cite{Schmitt1993} discusses this in more detail.

 \begin{quote}
"Coinjection is a partially-defined function whose restriction to where it is defined is a bijection; an example is $f: {1,2,3,4} \rightarrow {7,8}$ with $f(1) = 8$, $f(3) = 7$, and $f(2)$, $f(4)$ undefined). Intuitively, these are combinatorial structures with a notion of restriction on a subset of their vertex set; one can restrict a graph to a subset of its vertices by considering only the edges connected to this subset (usually known as the \textit{induced subgraph})". 

  \cite{Pang2014}
  \end{quote}

\subsubsection{OV circuit}

The OV circuit operates on the candidate values and combines them to produce a single output. Therefore it belongs to the product circuit.

\cite{Elhage2021} discusses the behavior of the OV circuit in terms of eigenvalues of the OV matrix. It was observed that there's a strong indication that if the matrix has positive eigenvalues, it is likely copying. Since the eigenvalues are real and positive we can conclude that it is a symmetric positive definite matrix.

A symmetric positive definite matrix, since it is a self-adjoint operator, is equal to its own conjugate transpose, i.e.

$$ A = A^T $$

Self-adjointness can be thought of as enforcing conservation of energy laws \cite{Ibragimov2011}. If we interpret the QK circuit as branching, we can interpret the the OV circuit as combining in the sense of \cite{Pang2014}. 

As such it can be thought of as the antipode \cite{Li2000} or intertwining operator.

In the context of Markov chains, \textbf{positive definite matrices can be understood as modeling interactions between Markov chains} \cite{OConnell2019}.

\subsection{Transformer learning mechanism}

The most important property of the residual stream is its basis independence, while the most important property of the attention mechanism is its symmetry and positive definiteness.

 \textbf{In our interpretation of transformers, the transformer model learns by enforcing Hopf coherence between the residual stream (unit response path) and the attention path (convolution path)}. 

\cite{Elhage2021} describes the attention mechanism as "sum where every term corresponds to an end-to-end path". This is fundamentally convolution as it can be considered a sum of all impulse responses \cite{Cheever2022}.

In our model learning works as follows:

\begin{enumerate}
	\item attention head receives a input
	\item attention head calculates output from both the unit response path and convolution path
	\item if the outputs match, the token gets copied
	\item if the outputs differ, the transformer propagates the difference (error) backward along the convolution path and unit impulse path, distributes the error between the QK and OV mechanisms and updates the residual stream (trace)
	\item by repeating this coproduct-product process (squaring) with different inputs, the transformer learns a stationary Markov chain distribution in its natural basis \cite{Diaconis2012}
\end{enumerate}

This mechanism of calculating the gradient of the loss function provides a better alternative to automatic differentiation since the error is calculated within the single layer and there's no explicit backward pass.

The details of this mechanism will be discussed in future work, however the intuition is that it is related to Laplace transform and Tannaka--Krein duality. 

Laplace transform has been observed to arise out of the interaction between convolution and shuffle product of combinatorial Hopf algebras \cite{Rutten2019}.

Tannaka--Krein, as a non-commutative generalization of Pontryagin duality, then obviates the relationship to Laplace transform.

\cite{Pavlovic2001} also explores the Laplace transform in the context of coinduction.

Another way of looking at it is via conjectured relationship between the induced representation (QK) and the intertwining operator (OV) on one end and the trace (residual stream) on the other \cite{Arthur2008}.

Dynamic programming provides a simplified, if not simplistic, model of certain elements of transformer models that is nonetheless rather useful. The main difference is that most aspects of dynamic programming algorithms are, rather ironically, quite static. In comparison, the transformer prunes its recurrences when combining a solution from subsolutions.

The interpretation in the context of DP also explains the "phase change" of transformer models. When the DP algorithm first starts filling out the grid, a lot of fields will be updated over a short period of time. Gradually, the frequency of the updates will decrease and the previous subsolutions will just be copied over.

\subsection{Attention head composition}

Zero layer attention only model learns bigram statistics. Since a bigram is just a Markov chain, these can be thought of as Hopf algebras without induced representation and intertwining operators. As a result, squaring the Hopf algebra similarly to \cite{Diaconis2012} makes the model learn a bigram. Chapter 5 of \cite{Majid1995} discusses how Hopf algebras can be used to model Markov chains. 

Ironically, in our model, the one layer and two layer attention models are somewhat similar. 

One layer attention only model then learns an ensemble of bigram and skip-gram and the two layer attention head then learns induction heads.

The exact mechanism how this works will be explored in future work, however the intuition is that since the shuffle product allows us to generate possible interleavings and ways of combining them, the chain "A ... B C" can be understood as a sum over all the coproducts which start with "A" and end with "B C".

\cite{EbrahimiFard2019} discusses these matters in terms of gap-insertion operad of non-crossing partitions.

\section{Conclusion and future work}

To summarize this paper, combinatorial Hopf algebras provides a rich algebraic infrastructure to interpret transformer models. This algebra has the surprising property of being able to calculate eigenvalues by repeated squaring.

By interpreting transformers as Hopf algebras, we arrive at a new understanding of the transformer learning mechanism as enforcement of Hopf coherence.

This mechanism boils down to enforcing coherence between the unit impulse path and the convolution path via Tannaka--Krein duality. As such, the transformer model can be understood as a linear time-invariant system.

In future work, we will discuss Hopf coherence in more detail. 

\bibliographystyle{apalike}
\bibliography{references}

\end{document}